\documentclass[sigconf]{acmart}
\usepackage{enumitem}
\usepackage{epstopdf}
\usepackage{multirow}
\usepackage{balance}
\usepackage[ruled,linesnumbered]{algorithm2e}
\usepackage{caption}
\usepackage{subcaption}
\usepackage{graphicx}
\usepackage{float}
\usepackage{adjustbox}
\usepackage{wrapfig,lipsum}
\usepackage{booktabs,etoolbox}
\usepackage{amsmath,amsfonts,amsthm,bm}
\usepackage{layouts}
\usepackage{cleveref}
\usepackage{tablefootnote}
\usepackage{makecell}
\usepackage{upgreek}
\usepackage{mathtools, stmaryrd}
\usepackage{algpseudocode}
\usepackage{xparse}
\DeclarePairedDelimiterX{\Iintv}[1]{\llbracket}{\rrbracket}{\iintvargs{#1}}
\NewDocumentCommand{\iintvargs}{>{\SplitArgument{1}{,}}m}
{\iintvargsaux#1} %
\NewDocumentCommand{\iintvargsaux}{mm} {#1\mkern1.5mu..\mkern1.5mu#2}

\DeclareMathOperator*{\argmin}{arg\,min}
\AtBeginDocument{%
  }

\usepackage{color}
\definecolor{darkgreen}{rgb}{0,0.5,0}
\definecolor{purple}{rgb}{1,0,1}

\def\ModelAbbr{OPTICAL}

\copyrightyear{2023} 
\acmYear{2023} 
\setcopyright{acmcopyright}\acmConference[WSDM '23]{Proceedings of the Sixteenth ACM International Conference on Web Search and Data Mining}{February 27-March 3, 2023}{Singapore, Singapore}
\acmBooktitle{Proceedings of the Sixteenth ACM International Conference on Web Search and Data Mining (WSDM '23), February 27-March 3, 2023, Singapore, Singapore}
\acmPrice{15.00}
\acmDOI{10.1145/3539597.3570468}
\acmISBN{978-1-4503-9407-9/23/02}

\begin{document}
\title{Improving Cross-lingual Information Retrieval on Low-Resource Languages via Optimal Transport Distillation}


\author{Zhiqi Huang}
\affiliation{%
  \institution{University of Massachusetts Amherst}
  \city{Amherst}
  \state{MA}
  \country{USA}
}
\email{zhiqihuang@cs.umass.edu}

\author{Puxuan Yu}
\affiliation{%
  \institution{University of Massachusetts Amherst}
  \city{Amherst}
  \state{MA}
  \country{USA}
}
\email{pxyu@cs.umass.edu}

\author{James Allan}
\affiliation{%
  \institution{University of Massachusetts Amherst}
  \city{Amherst}
  \state{MA}
  \country{USA}
}
\email{allan@cs.umass.edu}







\renewcommand{\shortauthors}{Zhiqi Huang, Puxuan Yu, \& James Allan}

\begin{abstract}
Benefiting from transformer-based pre-trained language models, neural ranking models have made significant progress. More recently, the advent of multilingual pre-trained language models provides great support for designing neural cross-lingual retrieval models. However, due to unbalanced pre-training data in different languages, multilingual language models have already shown a performance gap between high and low-resource languages in many downstream tasks. And cross-lingual retrieval models built on such pre-trained models can inherit language bias, leading to suboptimal result for low-resource languages.
Moreover, unlike the English-to-English retrieval task, where large-scale training collections for document ranking such as MS MARCO are available, the lack of cross-lingual retrieval data for low-resource language makes it more challenging for training cross-lingual retrieval models.
In this work, we propose \ModelAbbr{}: \underline{Op}timal \underline{T}ransport dist\underline{i}llation for low-resource \underline{C}ross-lingual information retriev\underline{al}. To transfer a model from high to low resource languages, \ModelAbbr{} forms the cross-lingual token alignment task as an optimal transport problem to learn from a well-trained monolingual retrieval model. By separating the cross-lingual knowledge from knowledge of query document matching, \ModelAbbr{} only needs \textbf{bitext data} for distillation training, which is more feasible for low-resource languages. Experimental results show that, with minimal training data, \ModelAbbr{} significantly outperforms strong baselines on low-resource languages, including neural machine translation.
\end{abstract}

\ccsdesc[500]{Information systems~Information retrieval}
\ccsdesc[500]{Information systems~Multilingual and cross-lingual retrieval}
\ccsdesc[300]{Information systems~Retrieval models and ranking}

\keywords{Cross-lingual information retrieval; Low-resource language; Knowledge distillation}

\maketitle
\section{Introduction}  \label{sec:introduction}
In the Cross-Lingual Information Retrieval (CLIR) task, the user submits the query in one language, and the systems respond by retrieving documents in another language. 
Different from the monolingual retrieval model, in addition to the ranking component,
the CLIR model needs an extra component of language translation to map the vocabulary of the query language to that of the documents’ language.  Therefore, the performance of a CLIR model depends on both the knowledge of query document matching and the ability to bridge the translation gap between query and document languages.

Pre-trained Transformer-based language models, such as BERT~\cite{devlin-etal-2019-bert}, have shown promising performance gains for monolingual information retrieval. This success is mainly due to two key factors: (i) The unsupervised pre-training of context-aware transformer architectures with an enormous number of parameters over large corpora. (ii) The fine-tuning for the downstream learning-to-rank task with a relatively large collection of relevance judgments, such as the MS MARCO passage ranking dataset~\cite{nguyen2016ms}. 
The multilingual versions of pre-trained Transformer-based language models, such as mBERT~\cite{devlin-etal-2019-bert} and XLM-R~\cite{conneau-etal-2020-unsupervised}, provide the possibility of jointly learning many languages with the same model. 
Because tokens in different languages are projected into the same space, fine-tuning these models with a cross-lingual retrieval dataset, similar to the monolingual setting, enables cross-language information retrieval. 

However, both factors leading to the success of monolingual information retrieval have defects in the cross-lingual setting. First, due to the unbalanced pre-training data in different languages, multilingual pre-trained models have already shown a performance gap between high and low-resource languages in many downstream tasks~\cite{wu2020all, wang-etal-2020-extending}. Cross-lingual retrieval models built on such pre-trained models can inherit the language bias, leading to suboptimal results for low-resource languages. 
Second, compared with the English-to-English retrieval task, the lack of cross-lingual IR training data with reliable relevance judgment, i.e., human relevance judgments, especially for low-resource languages, makes it more challenging to learn cross-lingual retrieval models.

Studies have attempted to address the data scarcity problem in CLIR. \citet{sasaki2018cross} proposed a large cross-lingual retrieval collection, WikiCLIR, based on the linked foreign language articles from Wikipedia pages. Because the Wikipedia articles in a specific language are edited mainly by native speakers, the cross-lingual contents in WikiCLIR are of high quality. But the relevant judgments are synthetically generated based on mutual links across pages. 
On the other hand, \citet{bonifacio2021mmarco} built a multilingual passage ranking dataset, mMARCO, by translating the queries and passages in MS MARCO into the target language using the neural machine translation (NMT) models. Because MS MARCO is generated from query log, the relevant judgments in mMARCO are more credible than WikiCLIR. Still, the automatically generated cross-lingual contents created by NMT models are not comparable to human writers, especially for resource-lean languages.

In this work, we present \ModelAbbr{}, a novel Optimal Transport-based knowledge distillation framework for low-resource CLIR task. 
From the modeling perspective, our approach separates the learning of query document matching from the learning of cross-lingual vocabulary mapping. More specifically, starting from the multilingual pre-trained encoder, we first train a bi-encoder English-to-English retrieval model, similar to the ColBERT architecture~\cite{khattab2020colbert}, as the \textit{teacher model}. This model can take full advantage of the MS MARCO triples to learn the knowledge for query document matching. 
Suppose the CLIR task is to search English documents with non-English queries. To devise a complete \textit{student model} for this CLIR task, we reuse the teacher model's document encoder, and train a new student query encoder. Given parallel queries, the non-English token representations generated by the student's query encoder are similar to the English token representations generated by the teacher’s query encoder.
In this step, the student model distills the retrieval knowledge from the teacher model in a cross-lingual setup. We form the distillation training as an optimal transport problem where the cost matrix is the cross-lingual token cosine distance, and the optimal transportation plan serves as a soft token alignment. The loss is then defined as the Frobenius inner product of the transportation plan and the cost matrix.
Because the teacher model already learns the knowledge of query-document matching and the distillation training only needs to focus on the translation knowledge as a general textual encoder, we can use bitext data to train the student query encoder, which is more feasible for low resource languages.

We performed extensive experiments on seven language pairs for CLIR training and evaluation, including four low-resource languages from diverse linguistic families and three medium or high-resource languages as a comparison. 
In terms of mean average precision (MAP), our proposed method significantly outperforms several strong baseline methods on low-resource languages, including a 13.7\% improvement over a method based on neural machine translation.
Further analysis demonstrates that the knowledge distillation step in \ModelAbbr{} is an effective and data-efficient method to transfer the knowledge of retrieval from monolingual into cross-lingual settings.

\section{Related Work}  \label{sec:related-work}

\subsection{Neural Matching Models for CLIR}
There are two sub-tasks within CLIR: translation and query-document matching. One approach is to translate the query into the language of the document collection and then apply a monolingual matching model to determine relevance. The translation could be accomplished by statistical machine translation (SMT)~\cite{bonab2020training} or neural machine translation (NMT)~\cite{saleh-pecina-2020-document}. While this two-step approach of translate-then-retrieve is popular, the emergence of bilingual word representation~\cite{vulic2015monolingual} and multilingual pre-trained language models~\cite{devlin-etal-2019-bert, conneau-etal-2020-unsupervised} create the opportunity to skip the translation step and match the query and document in different languages in a shared representation space. Because the token representation generated by multilingual pre-trained language models is contextualized based on other tokens in the sequence, once finetuned, they are effective across various tasks, including CLIR~\cite{litschko2018unsupervised, yu2021cross, Litschko2018, yu2020study}.

Learning neural matching models requires cross-lingual relevance knowledge for effective matching, whether based on multilingual word embeddings or pre-trained language models. 
The ideal data used for training CLIR models is expected to have both retrieval knowledge (query-document relevance) and translation knowledge (semantics across languages).
\citet{sasaki2018cross} constructed a large-scale, weakly supervised CLIR collection based on the linked foreign language articles from Wikipedia pages. They use the first sentence of a Wikipedia page as the query and all linked foreign language articles as the relevant documents. \citet{zhao2019weakly} leveraged parallel sentence data to create weakly supervised relevant judgments. They use a sentence from one language as the document and randomly translate a word from that sentence as the query. These CLIR collections contain the correct translation knowledge, but their retrieval knowledge is synthetically generated. 
On the other hand, some CLIR collections are created by translating a query from a commercial search engine into the target languages using NMT models~\cite{li-etal-2022-learning-cross,bonifacio2021mmarco}. The relevance judgments are more credible in these collections since they are extracted from the query log. However, their translation knowledge is compromised, especially in low-resource languages where the NMT models are not performing well. Besides cross-lingual relevance data, external knowledge, such as word-level translation knowledge, is also utilized to close the language gap in CLIR. 
\citet{bonab2020training} showed that dictionary-oriented word embeddings can improve the performance of a DRMM model~\cite{guo2016deep} when fine-tuned with relevance data. \citet{huang2021mixed} proposed a mixed attention transformer architecture to learn jointly from relevance judgments and word-level translation knowledge. 

Different from these approaches, we build a CLIR model with both translation knowledge and retrieval knowledge in two steps. First, a teacher model learns the retrieval knowledge via a monolingual retrieval collection. That knowledge is then transferred from the teacher model to a student model through knowledge distillation using bitext data.

\subsection{Knowledge Distillation}
Proposed by \citet{hinton2015distilling}, knowledge distillation is a method to train a model, called the student, using valuable information provided by the output of another model, called the teacher. This way, the teacher model's knowledge can be transferred into the student model. The idea of knowledge distillation is wildly used in the field of computer vision~\cite{xie2020self, yuan2019ckd, Lin_2022_CVPR}, natural language processing~\cite{sanh2019distilbert, reimers2020making} and information retrieval~\cite{lu2020twinbert, hofstatter2020improving, li-etal-2022-learning-cross}.
Our method is also an extension of knowledge distillation. A typical framework for knowledge distillation relies on a teacher model to directly generate a target distribution~\cite{gou2021knowledge,ma2022deep}.
Our method is different since the target distribution is determined by the optimal transport plan based on the cost matrix estimated by a teacher model. 

For the CLIR task, \citet{li-etal-2022-learning-cross} proposed a cross-lingual distillation framework to build a CLIR model from an English retriever. 
They use relevance score distillation. To compute the score in both teacher and student models, their method still requires retrieval-based data (i.e., parallel query) in the target language. 
Unlike their approach, we use the distance of token representation as a distillation signal. Thus, data with cross-lingual knowledge is adequate to train the student model in our approach. And such data (i.e., bitext data) is much easier to acquire.

    
    
    

\subsection{Optimal Transport}
Optimal Transport (OT) is a theory that studies the optimal allocation of resources between two probability distributions. Given a cost function, in order to compute the optimal transportation plan between two distributions, \citet{cuturi2013sinkhorn} leveraged the Sinkhorn's matrix scaling algorithm with an entropic regularization term to create a fast OT solver. 
\citet{pmlr-v115-xie20b} improved the Sinkhorn algorithm based on the proximal point method and constructed an IPOT solver which is robust to the parameter selection. In our method, we consider the estimation of the distance between two sets of token representations as an OT problem and adapt the IPOT solver to compute the optimal transportation plan.

The definition of OT has been applied to many areas, such as domain adaptation~\cite{flamary2016optimal}, generative models~\cite{salimans2018improving, chen2019improving} and self-supervision learning~\cite{wu2021data, lu2022faculty}. For tasks involving cross-lingual settings, \citet{nguyen2022improving} employed OT distance as a part of the loss function in a knowledge distillation framework for improving the cross-lingual summarization. \citet{alqahtani2021using} incorporated OT as an alignment objective to improve the multilingual word representations. In this work, we explore transferring the retrieval knowledge in a cross-lingual setting via OT.

\section{Methodology}  \label{sec:methodology}
Our goal is to incorporate the knowledge of query document matching from a well-learned monolingual retrieval model into a multilingual transformer-based retrieval architecture, such that it is capable of generating contextual representations under the cross-lingual setting and thus performing query document matching in different languages.
In this section, we first introduce the monolingual retrieval model as the teacher model. Then we present the optimal transport knowledge distillation framework and the training of the student model. Finally, we combine the components from both the teacher and student models into a CLIR model. Due to space limitations, we focus on the CLIR case of searching an English collection with a non-English query as an example to describe our method.

\subsection{Teacher Model}
Khattab and Zaharia~\cite{khattab2020colbert} discovered that instead of requiring both query and document to be present simultaneously at the beginning of the encoding process, the matching mechanism could be deferred until the contextualized representation computation is complete. Therefore, they proposed ColBERT, a bi-encoder retrieval model that first encodes the query and document separately and then scores based on late interaction between the token representations.
Because query and document encoders are relatively independent of each other, the ColBERT architecture in CLIR task has the potential to expand to a new language only for one type of encoder (i.e., query encoder) while leaving another one intact.  
For instance, when dealing with CLIR tasks between high and low resource languages, such as searching an English collection with a non-English query, adopting the bi-encoder separation design enables the model to reuse one component that is already well-trained using high resource monolingual retrieval data (i.e., document encoder). 
To leverage such modeling flexibility, we choose the teacher model in \ModelAbbr{} to follow the similar model architecture of ColBERT.

The teacher model $M$ contains two components: query encoder $E_{M_q}$ and document encoder $E_{M_d}$. Given a query $q$ and a candidate document $d$, the score of matching between $q$ and $d$, $S_{q,d}$, is then computed as the:
\begin{align}
\label{eq:score}
   S_{q,d} = \sum\limits_{i=1}^{|q|} \max\limits_{j=1}^{|d|} \ E_{M_q}(q_i) \cdot E^T_{M_d}(d_j)
\end{align}
where $E_{M_q}(q_i)$ is the $i$-th token representation of the query and $E_{M_d}(d_j)$ is the $j$-th token representation of the document. Because the output token representations from the encoders are normalized to unit length, the dot product is equivalent to cosine similarity. The scoring function applies the \textit{maxsim} operation on each query token to softly search against all document tokens to find the best token that reflects its context and then sums over all the query tokens. 
The primary goal of the teacher model is to provide knowledge of query document matching regardless of the language. Therefore, we select the dataset in the language that has the highest retrieval data quality to train the teacher model. Considering the quality and scale, we choose the English MS MARCO passage ranking dataset for teacher model training. Similar to ColBERT, we prepend special tokens \texttt{[Q]} and \texttt{[D]} to query and passage tokenization, respectively, and expand the query to a fixed length $L$ using the \texttt{[MASK]} token. Unlike ColBERT, we initialize the teacher model using a multilingual pre-trained model, mBERT instead of BERT. Since mBERT has a larger vocabulary that covers a more diverse set of languages, the student can benefit from the multilingual pre-trained token representation. 

\subsection{Optimal Transport Knowledge Distillation}
The student model shares the same architecture as the teacher. Note that if the document collection for the CLIR task remains in English, then the document encoder of the student model $E_{S_d}$ can be a copy of the teacher’s document encoder, $E_{M_d}$. Here, we focus on the design of the query encoder of the student model, $E_{S_q}$, which handles non-English queries.
Assume that $q$ is an English query and $\hat{q}$ is a non-English parallel query. The token representation of $q$ generated by $E_{M_q}$ contains rich knowledge for query document matching. If we could let $E_{S_q}$ ``behave'' like $E_{M_q}$, namely, if the output of $E_{S_q}$ with $\hat{q}$ is close to the output of $E_{M_q}$ with $q$, 
then the token representations generated by $E_{S_q}$ can have a similar retrieval performance to the teacher model.
Therefore, the training objective of knowledge distillation is to reduce the distance between the outputs of the teacher and student query encoders given parallel inputs (sentences). Next, we define the distance from $\hat{q}$ to $q$ in the vector space of $E_{S_q}$ and $E_{M_q}$. 
Because words that are translations of each other from different languages tend to have a smaller distance in the hyper-space after the multilingual pre-training step, we initialize the student query encoder's parameters weights using the same multilingual pre-trained language model as the teachers. Suppose the byte-pair encoding (BPE) tokenizer tokenizes $q$ into $L_q$ tokens and $\hat{q}$ into $L_{\hat{q}}$ tokens, we expand them to the same length $L$ by appending \texttt{[MASK]} tokens. After encoding by $E_{S_q}$ and $E_{M_q}$, $\hat{q}$ and $q$ are represented by a bag of vectors of size $L$, respectively. We define the distance from $\hat{q}$ to $q$ as follows:
\begin{align}
\label{eq:dis}
    D(\hat{q}, q) = \argmin\limits_{f_{\Iintv{1,L} \rightarrowtail \Iintv{1,L}}} \frac{1}{L} \sum\limits_{i=1}^{L} 1 - E_{S_q}(\hat{q}_i) \cdot E^{T}_{M_q}(q_{f(i)}) 
\end{align}

\SetAlFnt{\small}
\setlength{\textfloatsep}{5pt}
\begin{algorithm}[t]
\caption{IPOT algorithm}
\label{alg:ipot}
\SetKwInput{KwInput}{Input}                
\SetKwInput{KwOutput}{Output}              
\SetKwFunction{FMain}{IPOT}
\SetKwProg{Fn}{Function}{:}{\KwRet $\bm{\widetilde{\upgamma}}$}
\DontPrintSemicolon
\KwInput{Probability mass function of source and target $\{\mu_s, \mu_t$\}, cost matrix $\mathbf{C}$ and step size $\beta$.}
\KwOutput{Approximated OT matrix $\bm{\widetilde{\upgamma}}$}
\Fn{\FMain{$\mu_s, \mu_t, \mathbf{C}, \beta$}}{
    $\bm{b} \leftarrow \frac{1}{L}\mathbf{1}_L$, $\bm{\upgamma}^{(1)} \leftarrow \mathbf{1}\mathbf{1}^T$\\
    $\mathbf{G} \leftarrow \exp(\mathbf{C} / \beta)$ \\
    \For{$t = 1, 2, 3 \ldots N$}{
        $\mathbf{Q} \leftarrow \bm{\upgamma}^{(t)} \odot \mathbf{G} $ \tcp*{Hadamard product}
        \For(\tcp*[f]{Set $K=1$ in practice}){$k = 1, \ldots K$}{$\bm{a} \leftarrow  \frac{\mu_s}{\mathbf{Q} \, \bm{b}}$, $\bm{b} \leftarrow  \frac{\mu_t}{\mathbf{Q}^T \bm{a}}$}
        $\bm{\upgamma}^{(t+1)} = \mathrm{diag}(\bm{a}) \mathbf{Q} \mathrm{diag}(\bm{b})$
    }
    $\bm{\widetilde{\upgamma}} \leftarrow \bm{\upgamma}^{(N+1)}$
}
\end{algorithm}

\begin{figure*}[t]
    \captionsetup[subfigure]{font=footnotesize,labelfont=footnotesize}
    \begin{subfigure}[t]{0.85\linewidth}
        \centering
        \includegraphics[width=\linewidth]{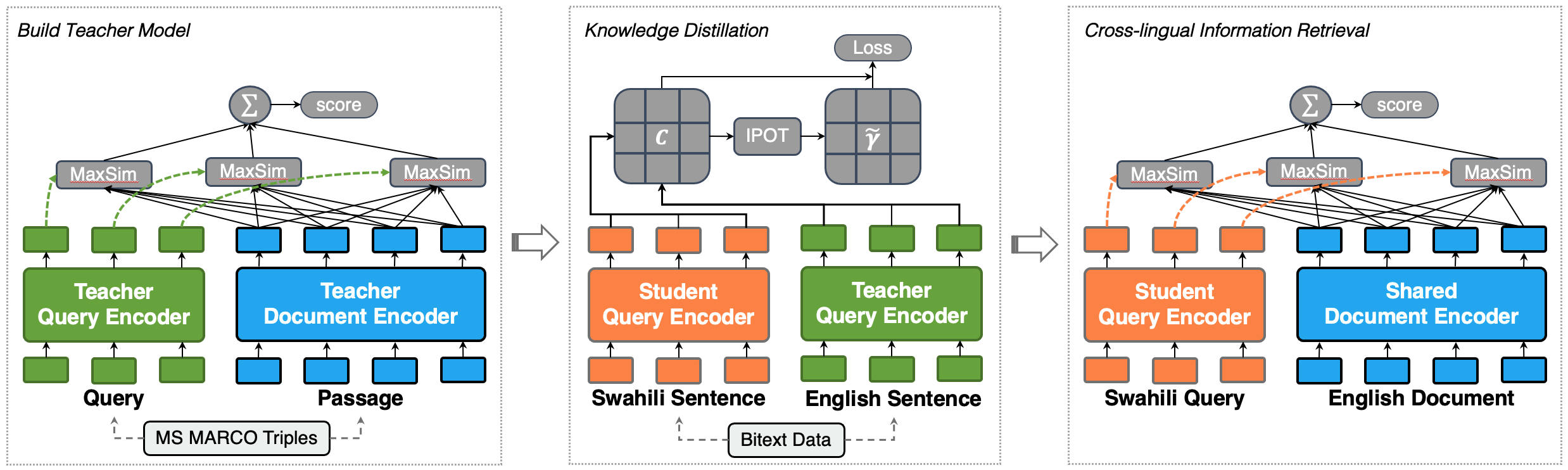}
    \end{subfigure}
\vspace{-5pt}
\caption{Model building pipeline for \ModelAbbr{}. This figure is based on CLIR task between Swahili query and English documents.}
\vspace{-5pt}
\label{fig:pipeline}
\end{figure*}


\noindent where $f$ is a bijective (one-to-one correspondence) function which maps the token index from $\hat{q}$ to $q$. Intuitively, the distance definition is equivalent to finding a token mapping from $\hat{q}$ to $q$ that minimizes the average of cosine distance among $L$ token pairs.
Despite the same semantics of $\hat{q}$ and $q$ as a whole, different languages have different token arrangements. When $L$ increases, using brute force to find the mapping $f$ becomes computationally intractable. Therefore, we approximate the calculation of $D(\hat{q}, q)$ as an optimal transport problem. First, we assign equal mass to the tokens in $\hat{q}$ and $q$ by defining a uniform source probability distribution, $\mu_s$, on $\hat{q}$ and a uniform target probability distribution, $\mu_t$, on $q$: 
$\mu_s (i)= \frac{1}{L} $ and $\mu_t (j) = \frac{1}{L}$ where $1\leq i,j \leq L$.

The set of transportation plans between these two distributions is then the set of doubly stochastic matrices $\mathcal{P}$ defined as
\begin{align}
    \mathcal{P} = \{ \bm{\upgamma} \in (\mathbb{R}^+)^{L\times L} \ | \ \bm{\upgamma} \mathbf{1}_{L} = \mu_s , \ \bm{\upgamma}^T \mathbf{1}_{L} = \mu_t\}
\end{align}
where $\mathbf{1}_{L}$ is a $L$-dimensional vector of ones and $\bm{\upgamma}$ is called a transportation plan. We redefine distance between $\hat{q}$ and $q$ as a Wasserstein distance between distribution $\mu_s$ and $\mu_t$. Then the computation of such distance become an optimal transport (OT) problem:
\begin{align}
\bm{\upgamma}_0 = \argmin\limits_{\bm{\upgamma} \in \mathcal{P}} \bigl \langle \bm{\upgamma}, \mathbf{C} \bigr \rangle_{F}
\end{align}
where $\bigl \langle \cdot , \cdot \bigr \rangle_{F}$ is the Frobenius inner product, $\bm{\upgamma}_0$ is the optimal transportation plan (or OT matrix) and $\mathbf{C} \geq 0$ is a $L\times L$ cost function matrix of term $C(i,j)$, reflecting the ``energy'' needed to move a probability mass from $\hat{q}_i$ to $q_j$. In our setting, this cost is chosen as the cosine distance between two tokens:
\begin{align*}
    C(i,j) = 1 - E_{S_q}(\hat{q}_i) \cdot E^{T}_{M_q}(q_{j})
\end{align*}
In general, the linear programming solution to find $\bm{\upgamma}_0$ has a typical super $O(n^3)$ complexity that is still computationally intractable~\cite{cuturi2013sinkhorn}. To overcome such intractability, we employ the Inexact Proximal point method for Optimal Transport (IPOT)~\cite{pmlr-v115-xie20b} algorithm to compute the OT matrix.
Specifically, the IPOT algorithm iteratively solves the OT problem by adding a proximity metric term to the original OT definition. At step $t$, we have:
\begin{align*}
    \bm{\upgamma}^{(t+1)} = \argmin\limits_{\bm{\upgamma} \in \mathcal{P}} \Bigl\{ \bigl \langle \bm{\upgamma}, \mathbf{C} \bigr \rangle_{F} + \beta \cdot \mathcal{B}(\bm{\upgamma}, \bm{\upgamma}^{(t)}) \Bigr\}
\end{align*}
where $\mathcal{B}(\bm{\upgamma}, \bm{\upgamma}^{(t)}) = \sum^{L}_{i,j}\bm{\upgamma}_{ij} \log \frac{\bm{\upgamma}_{ij}}{\bm{\upgamma}^{(t)}_{ij}} - \sum^{L}_{i,j}\bm{\upgamma}_{ij} + \sum^{L}_{i,j}\bm{\upgamma}^{(t)}_{ij} $ is the Bregman divergence used as a proximity metric term to penalize the distance between the solution and the latest approximation. It provides a tractable iterative scheme toward the exact OT solution where the step size is controlled by $\beta$. The implementation details for IPOT are in Algorithm \ref{alg:ipot}. 
Using the approximated OT matrix, we define the loss of the distillation as the total transportation cost:
\begin{align}
    loss \vcentcolon = \bigl \langle \bm{\widetilde{\upgamma}}, \mathbf{C} \bigr \rangle_{F}
\end{align}
During training, we retain the teacher query encoder by removing its parameters from the computational graph and use the loss to update the student query encoder. Given each pair of $\hat{q}$ and $q$, minimizing the loss will lead the model to reduce the cosine distance between tokens according to the transportation plan. And because $E_{M_q}$ is fixed, the essence of knowledge distillation is to push non-English token representations generated by $E_{S_q}$ towards their corresponding English token representations generated by $E_{M_q}$. 
Moreover, though designed as the query encoder, the textual data of $\hat{q}$ and $q$ used for distillation do not have to be the query from a CLIR dataset. A group of parallel sentences with a broad vocabulary coverage is adequate to train the student query encoder. Compared to the CLIR data, which often require human relevant judgments, bitext data are easier to acquire, especially for low-resource languages. And any improvements regarding the knowledge of query document matching in monolingual retrieval performance (teacher model) can be easily transferred to the cross-lingual setting (student model) using bitext data with the \ModelAbbr{} framework. 

\subsection{Cross-lingual Query Document Matching}
In this section, we focus on a CLIR task of searching English documents using a non-English query to introduce the \ModelAbbr{} framework. The document encoder in the student model can be directly copied from the teacher model ($E_{S_d}$\textleftarrow$E_{M_d}$). At test time, the matching score of $\hat{q}$ and $d$ is calculated based on equation~(\ref{eq:score}) using $E_{S_q}$ and $E_{S_d}$.
A complete overview of the model building pipeline is shown in Figure \ref{fig:pipeline}. 
In fact, \ModelAbbr{} can be extended to different language settings in the CLIR task. For example, suppose the task requires searching non-English documents using an English query. In this case, the student model can reuse the teacher’s query encoder and train a non-English document encoder using knowledge distillation. More generally, if the query is in $X$ and the collection is in $Y$, where $X$ and $Y$ are both non-English languages, we can build the student model by training two knowledge distillations: $X$ to English for query encoder and $Y$ to English for document encoder.

\section{Experimental Setup}  
\label{sec:exp_setup}
\subsection{CLIR Settings and Dataset}
\textbf{CLIR Settings.} 
We focus on searching English collections with queries in low-resource languages. We consider four low-resource languages in our experiments: Swahili, Somali, Tagalog, and Marathi. According to linguistic classification\footnote{https://en.wikipedia.org/wiki/List\_of\_language\_families}, they belong to four different language families: Niger-Congo (Swahili), Afro-Asiatic (Somali), Austronesian (Tagalog), and Indo-European (Marathi). To fully evaluate the performance of the proposed method, we also include three medium to high-resource languages: Finnish, German and French.

\textbf{Evaluation data.}
We create a unified evaluation dataset for all language pairs considered in our experiments. The data are from the Cross-Language Evaluation Forum (CLEF) 2000-2003 campaign for bilingual ad-hoc retrieval tracks~\cite{braschler2002clef}. The query is a concatenation of the title and description fields of the topic files. In total, there are 151 queries from the CLEF C001 – C200 topic (queries with no relevant judgment are removed). The collection of English documents is the Los Angeles Times corpus comprised of 113k news articles. For Finnish, German, and French, their queries are provided by CLEF campaign. For low-resource languages, \citet{10.1145/3341981.3344236} provided
Somali and Swahili translations of English queries. And we hire bilingual human experts from Gengo\footnote{https://gengo.com} service to translate English queries into Tagalog and Marathi. 

\textbf{Retrieval training data.}
To guarantee a consistent performance of the teacher model on the monolingual retrieval task, we randomly sample a subset of 7 million triples from the MS MARCO passage ranking dataset for the training of the teacher model. The baselines, which involve synthesizing the CLIR dataset with different methods, all use the same subset of triples.

\textbf{Bitext training data}.
To support the cross-lingual knowledge distillation, we use the parallel sentences from the CCAligned dataset~\cite{el-kishky-etal-2020-ccaligned}. In section 5.2, for the main result table, the distillation is trained based on a random sample of up to 2 million parallel sentences for each language pair. In section 5.4, we study the effect of bitext data size on the performance of the student model.
Note that for Somali and Marathi, the total number of parallel sentences in CCAligned is less than 2 million. Thus, we use all the parallel sentences available for these two languages (360K for Somali and 750K for Marathi). Moreover, there are other parallel corpora for the languages studied in our experiments that could help us to create larger training data. We only use CCAligned to ensure the consistency of the data quality.

\begin{table*}[!t]
    \centering
    \captionsetup{width=\linewidth}
    \caption{Size of language data resource and OPUS-MT model performance.}
    \vspace{-8pt}
    \label{tab:nmt}
    \begin{adjustbox}{width=0.9\textwidth}
    \aboverulesep=0ex
    \belowrulesep=0ex
    \renewcommand{\arraystretch}{1.2}
    \begin{tabular}{lcccccccccccccc}
        \toprule
       \multirow{2}{*}[-6pt]{\textbf{NMT Models}} & \multicolumn{8}{c}{Low Resource Languages} & \multicolumn{6}{c}{Medium or High Resource Languages} \\
       \cmidrule(lr){2-9} \cmidrule(lr){10-15}
       & \multicolumn{2}{c}{Swahili (SW)} & \multicolumn{2}{c}{Somali (SO)} & \multicolumn{2}{c}{Tagalog (TL)} & \multicolumn{2}{c}{Marathi (MR)} & \multicolumn{2}{c}{Finnish (FI)} & \multicolumn{2}{c}{German (DE)} & \multicolumn{2}{c}{French (FR)}\\
        \midrule

        
        Train./eval. data available & \multicolumn{2}{c}{9M/386} & \multicolumn{2}{c}{0.8M/4} & \multicolumn{2}{c}{8M/2,500} & \multicolumn{2}{c}{5M/10,369} & \multicolumn{2}{c}{45M/10,690} & \multicolumn{2}{c}{86M/17,565} & \multicolumn{2}{c}{180M/12,681} \\

 
        Translation direction & EN-SW & SW-EN & EN-SO & SO-EN & EN-TL & TL-EN & EN-MR & MR-EN & EN-FI & FI-EN & DE-EN & EN-DE & EN-FR & FR-EN \\
        \cmidrule(lr){2-3} \cmidrule(lr){4-5} \cmidrule(lr){6-7} \cmidrule(lr){8-9} \cmidrule(lr){10-11} \cmidrule(lr){12-13} \cmidrule(lr){14-15} 
        BLEU scores & 26.0 & 31.3 & 16.0 & 23.6 & 26.5 & 35.0 & 18.2 & 29.8 & 40.4 & 50.9 & 47.3 & 55.4 & 50.5 & 57.5 \\
        \bottomrule
    \end{tabular}
    \end{adjustbox}
\vspace{-5pt}
\end{table*}

\begin{table*}[!t]
    \centering
    \captionsetup{width=\linewidth}
    \caption{First-Stage Retrieval Comparison. For Recall columns, the highest value is marked with bold text. Note that the first row is reported as an upper bound reference.}
    \vspace{-8pt}
    \label{tab:first-stage}
    \begin{adjustbox}{width=0.9\textwidth}
    \aboverulesep=0ex
    \belowrulesep=0ex
    \renewcommand{\arraystretch}{1.2}
    \begin{tabular}{lcccccccccccccc}
        \toprule
       \multirow{3}{*}[-6pt]{\textbf{\shortstack[l]{First-Stage\\Retrieval}}} & \multicolumn{8}{c}{Low Resource Languages} & \multicolumn{6}{c}{Medium or High Resource Languages} \\
       \cmidrule(lr){2-9} \cmidrule(lr){10-15}
       & \multicolumn{2}{c}{Swahili} & \multicolumn{2}{c}{Somali} & \multicolumn{2}{c}{Tagalog} & \multicolumn{2}{c}{Marathi} & \multicolumn{2}{c}{Finnish} & \multicolumn{2}{c}{German} & \multicolumn{2}{c}{French}\\
        \cmidrule(lr){2-3} \cmidrule(lr){4-5} \cmidrule(lr){6-7} \cmidrule(lr){8-9} \cmidrule(lr){10-11} \cmidrule(lr){12-13} \cmidrule(lr){14-15} 
        & MAP & Recall & MAP & Recall & MAP & Recall & MAP & Recall & MAP & Recall & MAP & Recall & MAP & Recall\\
        \midrule
        Human+BM25 & 0.4569 & 0.7621 & 0.4569 & 0.7621 & 0.4569 & 0.7621 & 0.4569 & 0.7621 & 0.4569 & 0.7621 & 0.4569 & 0.7621 & 0.4569 & 0.7621 \\
        \midrule
        SMT+BM25 & 0.2184 & 0.4359 & 0.1948 & \textbf{0.4254} & 0.1636 & 0.6195 & 0.1059 & 0.3289 & 0.3052 & 0.6049 & 0.3906 & 0.6946 & 0.4037 & 0.7541\\
        NMT+BM25 & 0.2135 & \textbf{0.4934} & 0.1466 & 0.3670 & 0.3501 & \textbf{0.6799} & 0.1795 & \textbf{0.4277} & 0.3753 & \textbf{0.7248} & 0.4087 & \textbf{0.7420} & 0.4315 & \textbf{0.7585}\\
        \bottomrule
    \end{tabular}
    \end{adjustbox}
\vspace{-8pt}
\end{table*}

\subsection{Implementation Details}
We initialize the ColBERT query and document encoder components in both teacher and student models using the multilingual pre-trained BERT model (mBERT, base, uncased). We set the max length of all query encoders at $L = 32$ and truncate the document at 180 tokens.
There are two model training tasks in our experiments. One is the \textbf{retrieval} training task. This is the main task of training the teacher model and the other neural baselines. Given a query, relevant passage, and non-relevant passage triplet, the models are trained using pairwise cross-entropy loss with a learning rate of $3\times 10^{-6}$ and a batch size of 64 for 200K iterations. 
The other training task is \textbf{knowledge distillation}. In this task we train the student model using bitext data. We set the step size to $\beta=0.5$ and number of iterations to $N=100$ for the IPOT algorithm. We use the cost of the optimal transportation as the loss and build a batch size of 256 with gradient accumulation. The student model is trained with a learning rate of $5\times 10^{-5}$ for 3 epochs of the available bitext data. All the experiments are implemented using Python 3 and PyTorch 1.8.1. The pre-trained model weights, including the neural machine translation models, are accessed from HuggingFace\footnote{https://huggingface.co/models}. 
Regarding to the NMT models used in some of our baseline methods, we use the off-the-shelf OPUS-MT~\cite{tiedemann-thottingal-2020-opus} from the Helsinki NLP group\footnote{\url{ https://huggingface.co/Helsinki-NLP}}. All the NMT models use the Marian-NMT~\cite{mariannmt} as the base architecture and are trained using the OPUS corpus\footnote{\url{ https://opus.nlpl.eu/}}. 

\textbf{Evaluation.} While we train models on passages for the retrieval task, our goal is to rank documents whose length is usually longer than 180 tokens. We split large documents into overlapping passages of fixed length with a stride of 90 tokens and compute the score for each query passage pair. Finally, we select a document’s maximum passage score as its document score~\cite{nair2022transfer}. For evaluating retrieval effectiveness, we follow prior work on the CLEF dataset~\cite{10.1145/3331184.3331324, bonab2020training} and report mean average precision (MAP) of the top 100 and precision of the top 10 (P@10) ranked documents. We determine statistical significance using the two-tailed paired \textit{t}-test with p-value less than 0.05 (i.e., 95\% confidence level).

\subsection{Compared Methods}
\subsubsection{First-Stage Retrieval.}
We employ a two-stage retrieval approach for addressing the
CLIR problem, where first we obtain an initial set of candidate documents using a lexical matching retrieval technique (e.g., Okapi BM25) and then re-rank the initial set of candidate documents using a neural re-ranker. We select Recall@100 as our primary evaluation metric for the first-stage retrieval to collect the most relevant documents.
We investigate different strategies for our initial ranking stage that we elaborate in the following:

\begin{itemize}[leftmargin=*,noitemsep,topsep=0pt]
\item \textbf{SMT+BM25}: We translate the query based on a statistical machine translation (SMT) method. More specifically, we first build a translation table from parallel corpora for each language pair using the GIZA++ toolkit. Then we select the top 10 translations from the translation table for each query term and apply Galago’s weighted \textit{\#combine} operator to form a translated query. Finally, we run BM25 with default parameters to retrieve documents.

\item \textbf{NMT+BM25}:
Neural machine translation models based on the encoder-decoder architecture are empirically better than SMT in terms of translation quality. Thus, we build this baseline by first translating the query into English using an NMT model. Then we run BM25 with default parameters to retrieve documents. 

\item \textbf{Human+BM25}:
For a comprehensive comparison, we also provide an empirical upper bound on the initial ranking stage. We use CLEF English query as the human translations and apply BM25 as the retrieval technique. 

\end{itemize}

\subsubsection{Neural Re-ranking.}
In the second retrieval stage, to provide the best initial rank list for neural re-ranking, we select the rank list from either SMT+BM25 or NMT+BM25 based on their Recall for each CLIR language pair. Including the proposed method, all the neural models rerank the top 100 documents of the initial rank list. We compare \ModelAbbr{} with the methods in the following:

\begin{itemize}[leftmargin=*,noitemsep,topsep=0pt]
\item \textbf{mColBERT}:
Because the encoders in the teacher model are based on a multilingual pre-trained language model, after training using English MS MARCO triples, we can directly run the teacher model on the CLIR evaluation data in a zero-shot setting.

\item \textbf{Code-Switch}:
There are data augmentation methods that can help the training of cross-lingual tasks. \citet{qin2020cosda} proposed a code-switching framework to transform the monolingual training data into data in mixed languages. \citet{bonab2020training} proposed a shuffling algorithm to inject and mix the translated terms into the query. We apply the code-switch method to the queries in MS MARCO triples. More specifically, we randomly switch 50\% of the English query words into their translations in the target language according to the Panlex dictionary and then train the ColBERT retrieval model using the code-switched data.

\item \textbf{Translate-Train}:
\citet{bonifacio2021mmarco} built a multilingual passage ranking dataset, mMARCO, by translating MS MARCO into target languages using the correspondent OPUS-MT models~\cite{tiedemann-thottingal-2020-opus}.
~\citet{nair2022transfer} showed that retrieval models trained using this synthetically generated CLIR dataset could outperform BM25 with query translation and the zero-shot neural approach in high resource languages. We adopt this method as another baseline. For languages that are not covered by mMARCO, we follow the same data generation procedure to build training data.

\item \textbf{Translate-Test}:
Like NMT-BM25, we can first let the NMT model translate the evaluation query into English and then perform English-to-English query document matching using a well-trained monolingual neural retrieval model (i.e., the teacher model). 

\item \textbf{Human+ColBERT}:
We also provide an empirical upper bound on the re-ranking stage.  We use human translations of the evaluation query and apply the teacher model to re-rank the top 100 documents from the rank lists generated by the Human+BM25. 
\end{itemize}

\begin{table*}[!t]
    \centering
    \captionsetup{width=\linewidth}
    \caption{A comparison of model performance. $\triangleright$ are reported as the upper bound reference. The highest value is marked with bold text. Statistically significant improvements are marked by $\dag$ (over Translate-Train) and $\ddag$ (over Translate-Test).}
    \vspace{-8pt}
    \label{tab:main}
    \begin{adjustbox}{width=0.98\textwidth}
    \aboverulesep=0ex
    \belowrulesep=0ex
    \renewcommand{\arraystretch}{1.2}
    \begin{tabular}{lcccccccccccccc}
        \toprule
       \multirow{3}{*}[-6pt]{\textbf{\shortstack[l]{Retrieval\\Methods}}} & \multicolumn{8}{c}{Low Resource Languages} & \multicolumn{6}{c}{Medium or High Resource Languages} \\
       \cmidrule(lr){2-9} \cmidrule(lr){10-15}
       & \multicolumn{2}{c}{Swahili} & \multicolumn{2}{c}{Somali} & \multicolumn{2}{c}{Tagalog} & \multicolumn{2}{c}{Marathi} & \multicolumn{2}{c}{Finnish} & \multicolumn{2}{c}{German} & \multicolumn{2}{c}{French}\\
        \cmidrule(lr){2-3} \cmidrule(lr){4-5} \cmidrule(lr){6-7} \cmidrule(lr){8-9} \cmidrule(lr){10-11} \cmidrule(lr){12-13} \cmidrule(lr){14-15} 
        & MAP & P@10 & MAP & P@10 & MAP & P@10 & MAP & P@10 & MAP & P@10 & MAP & P@10 & MAP & P@10 \\
        \midrule
        $\triangleright$ Human+BM25 & 0.4569 & 0.3940 & 0.4569 & 0.3940 & 0.4569 & 0.3940 & 0.4569 & 0.3940 & 	0.4569 & 0.3940 & 0.4569 & 0.3940 & 0.4569 & 0.3940\\
        \midrule
        SMT+BM25 & 0.2184 & 0.2152 & 0.1948 & 0.1865 & 0.1636 & 0.0934 & 0.1059 & 0.0984 & 0.3052 & 0.2821 & 0.3906 & 0.3437 & 0.4037 & 0.3772\\
        NMT+BM25 & 0.2135 & 0.2113 & 0.1466 & 0.1380 & 0.3501 & 0.3179 & 0.1795 & 0.1795 & 0.3753 & 0.3583 & 0.4087 & 0.3580 & 0.4315 & 0.3881 \\
        \midrule
        $\triangleright$ Human+ColBERT & 0.5019 & 0.4344 & 0.5019 & 0.4344 & 0.5019 & 0.4344 & 0.5019 & 0.4344 & 0.5019 & 0.4344 & 0.5019 & 0.4344 & 0.5019 & 0.4344 \\
        \midrule
        mColBERT & 0.1953 & 0.1795 & 0.1355 & 0.1212 & 0.3414 & 0.2960 & 0.1448 & 0.1556 & 0.3791 & 0.3272 & 0.4509 & 0.3807 & 0.4512 & 0.3868 \\
        Code-Switch & 0.2420 & 0.2258 & 0.1845 & 0.1682 & 0.3542 & 0.2934 & 0.1573 & 0.1662 & 0.3831 & 0.3404 & 0.4553 & 0.3827 & 0.4589 & 0.3993 \\
        Translate-Train & 0.2234 & 0.2185 & 0.1707 & 0.1649 & 0.3692 & 0.3252 & 0.1619 & 0.1722 & 0.4043 & 0.3576 & 0.4713 & 0.3967 & 0.4666 & 0.4020 \\
        Translate-Test & 0.2643 & 0.2530 & 0.2126 & 0.2086 & 0.3827 & 0.3339 & 0.2141 & 0.2258 & $\mathbf{0.4418}$ & $\mathbf{0.4024}$ & 0.4811 & $\mathbf{0.4080}$ & $\mathbf{0.4984}$ & $\mathbf{0.4318}$ \\
        \ModelAbbr{} & $\mathbf{0.3129}^{\dag\ddag}$ & $\mathbf{0.2901}^{\dag\ddag}$ & $\mathbf{0.2477}^{\dag\ddag}$ & $\mathbf{0.2365}^{\dag\ddag}$ & $\mathbf{0.4188}^{\dag\ddag}$ & $\mathbf{0.3623}^{\dag\ddag}$ & $\mathbf{0.2414}^{\dag\ddag}$ & $\mathbf{0.2384}^{\dag}$ & $0.4228$ & $0.3874^{\dag}$ & $\mathbf{0.4832}$ & $0.4067$ & $0.4764$ & $0.4119$ \\
        \bottomrule
    \end{tabular}
    \end{adjustbox}
\vspace{-8pt}
\end{table*}

\section{Experimental Results}  
\label{sec:exp_results}

\subsection{First-Stage Retrieval Comparison}
Table~\ref{tab:first-stage} shows the results of our first-stage retrieval methods. We can see that the NMT+BM25 approach outperforms SMT+BM25 in Recall@100 for all languages except Somali. Referring to the evaluation in Table~\ref{tab:nmt}, the failure of NMT+BM25 on Somali is mainly due to the poor translation quality. Moreover, human translation performs better than machine translation, and the margin is larger in low-resource than high-resource languages, indicating higher difficulty in building machine translation systems in low-resource languages. Based on recall, we choose the rank lists from NMT + BM25 for the reranking step for all languages except Somali, for which we use SMT+BM25 as the initial retrieval method.

\subsection{Re-ranking Comparison}
Table~\ref{tab:main} lists the evaluation results of both the first-stage retrieval methods and neural re-ranking models. For the zero-shot setting, we can see that mColBERT can improve the initial ranking on high-resource languages while failing on low-resource languages. 
Similar to other downstream tasks~\cite{wu2020all, wang-etal-2020-extending}, the CLIR model based on a multilingual pre-trained language model also inherits the language bias in the pre-training step, causing the performance gap between low and high resource languages in document ranking.
Using dictionary knowledge for cross-lingual data augentation, the Code-Switch method performs better than mColBERT. However, the word-level translation knowledge used in Code-Switch does not consider the context of the switched terms, which could cause the semantics of the code-switched data to diverge from the original one.
Comparing Code-Switch to Translate-Train, we can see that Translate-Train outperforms Code-Switch in high-resource languages. With the support of the NMT model, the Translate-Train method can generate better query translations in high-resource languages, which leads to a higher quality of CLIR training triples than the Code-Switch method. However, in low-resource languages Swahili and Somali, Translate-Train cannot consistently outperform Code-Switch. This is because the NMT model does not have enough training resources, and the generated query translations are of lower quality. 
Instead of building a CLIR dataset for model training, the Translate-Test method translates the query to English using an NMT model and then ranks the document based on a monolingual neural retrieval model. With the help of two neural models at test time (translation and document ranking), this two-step approach becomes the strongest baseline in our experiment.

Finally, we can see the supremacy of \ModelAbbr{} in low-resource languages. Our method consistently and significantly improves the first-stage retrieval results.  In low-resource languages, \ModelAbbr{} substantially outperforms all baselines, including the Translate-Test method. In high-resource languages, \ModelAbbr{} also achieves solid performance. It outperforms the Translate-Train method in all three languages. And it is a surprise to us that \ModelAbbr{} exceeds the Translate-Test on German in terms of MAP.
Moreover, the results of \ModelAbbr{} in Table~\ref{tab:main} are only based on a maximum of 2M parallel sentences (there are only 360K for Somali and 750K for Marathi). No cross-lingual relevance judgment is used in the distillation step, making \ModelAbbr{} data feasible and easy to build.
At the same time, we can see that using human translation with a neural ranking model (Human+ColBERT) still leads the CLIR setting with the same model architecture by a large margin in low-resource languages.

\subsection{Analysis of Knowledge Distillation}
To study how the student model behaves after distillation by the \ModelAbbr{} framework, we compare the token representations of the same query generated by different models.

\begin{figure}[t]
    \captionsetup[subfigure]{aboveskip=2pt,font=footnotesize,labelfont=footnotesize}
    \begin{subfigure}[t]{0.45\textwidth}
        \centering
        \includegraphics[width=\linewidth]{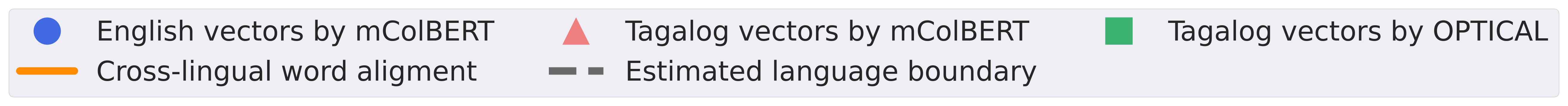}
        \vspace{-0.4cm}
    \end{subfigure}
    \begin{subfigure}[t]{0.22\textwidth}
        \centering
        \includegraphics[width=\linewidth]{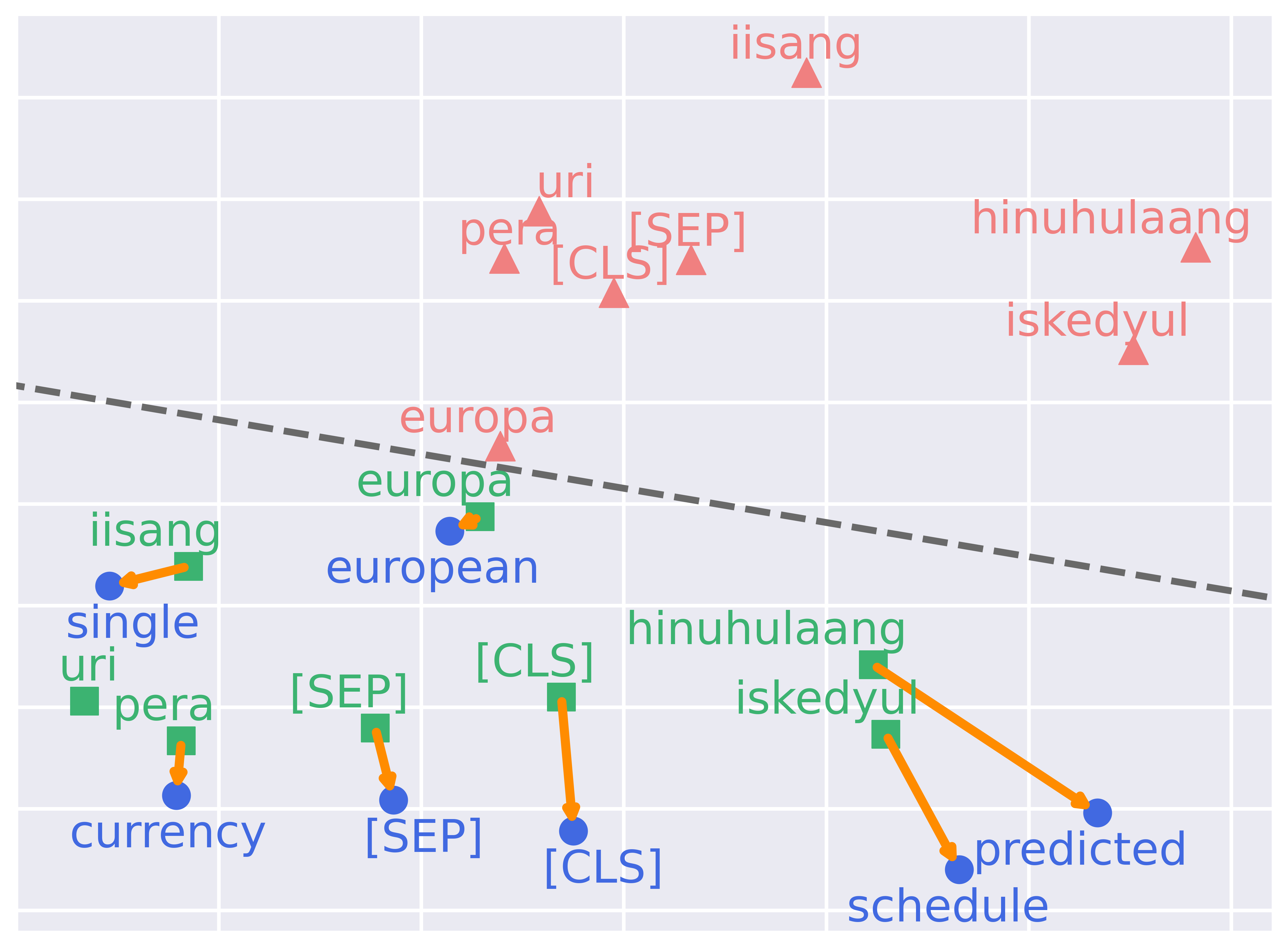}
        \caption{Example query comparison.}
        \label{fig:tsne-tl-example}
    \end{subfigure}
    \begin{subfigure}[t]{0.22\textwidth}
        \centering
        \includegraphics[width=\linewidth]{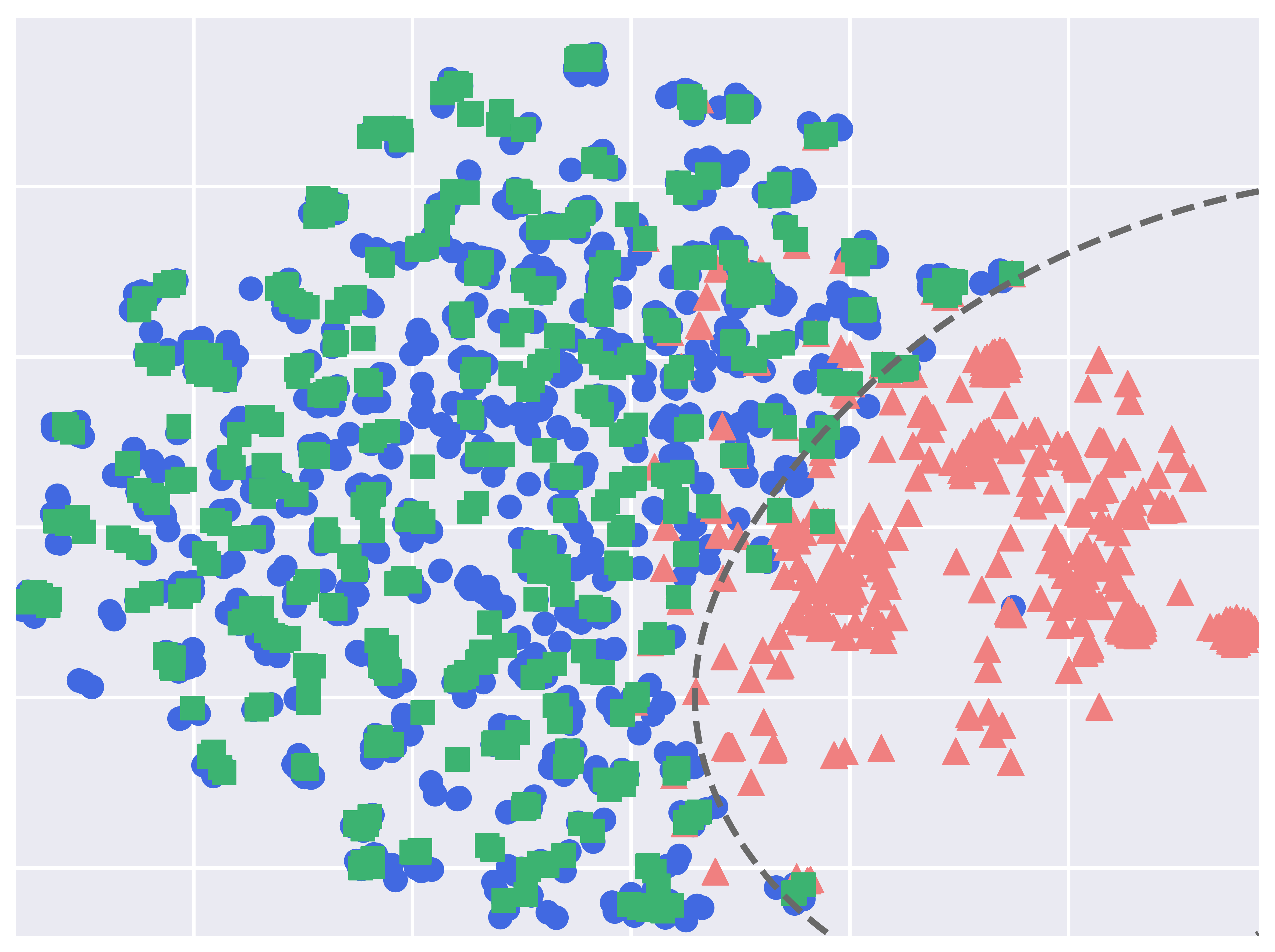}
        \caption{Overview of all test queries.}
        \label{fig:tsne-tl-overview}
        \vspace{5pt}
    \end{subfigure}
    \begin{subfigure}[t]{0.45\textwidth}
        \centering
        \includegraphics[width=\linewidth]{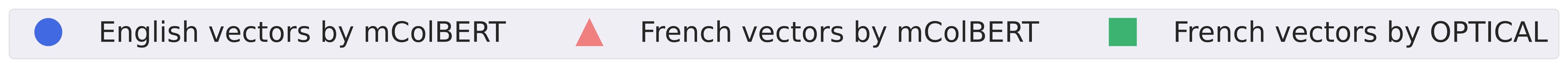}
        \vspace{-0.4cm}
    \end{subfigure}
    \begin{subfigure}[t]{0.22\textwidth}
        \centering
        \includegraphics[width=\linewidth]{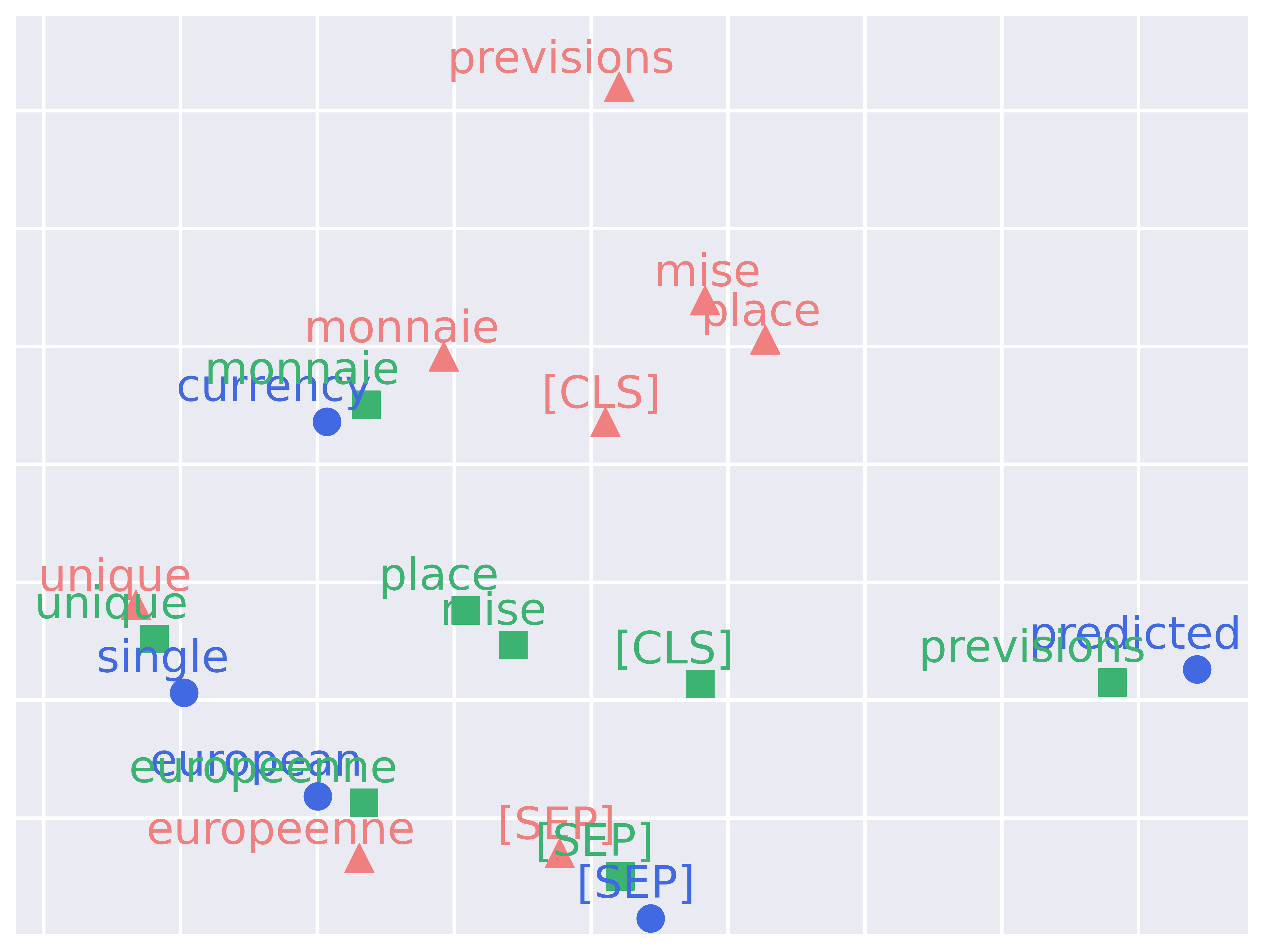}
        \caption{Example query comparison.}
        \label{fig:tsne-fr-example}
    \end{subfigure}
    \begin{subfigure}[t]{0.22\textwidth}
        \centering
        \includegraphics[width=\linewidth]{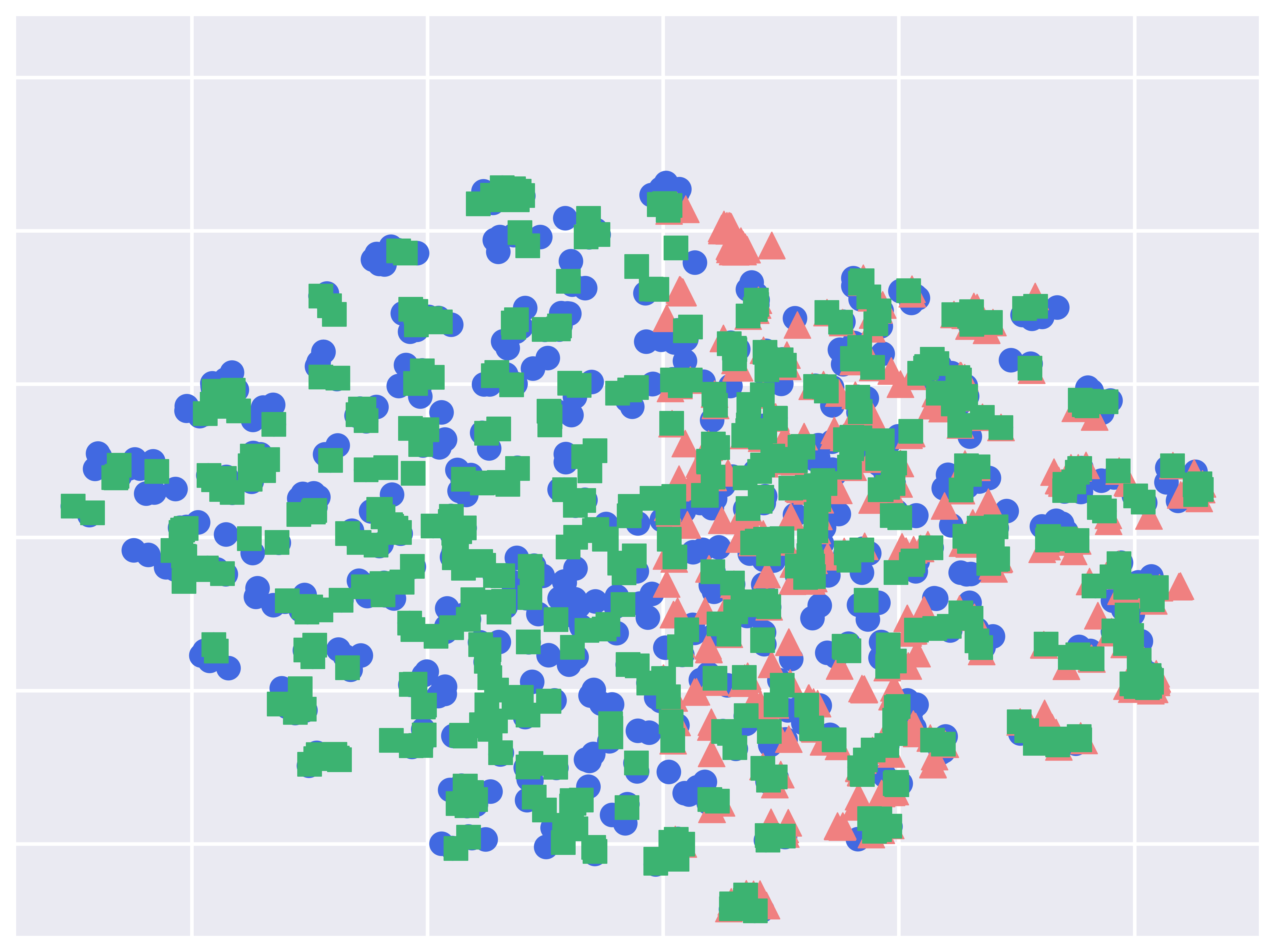}
        \caption{Overview of all test queries.}
        \label{fig:tsne-fr-overview}
    \end{subfigure}
\vspace{-8pt}
\caption{t-SNE visualisation of query tokens.}
\vspace{-6pt}
\label{fig:tsne}
\end{figure}

\textbf{Student query encoder in low-resource language.}
We consider three types of token representations. First, we encode the English query using the mColBERT query encoder. Note that the mColBERT query encoder is the same as the \ModelAbbr{} teacher encoder, which provides the knowledge to the student query encoder during the distillation.
Then we encode the same corresponding Tagalog query using both mColBERT and \ModelAbbr{} student query encoders. Finally, we use t-SNE~\cite{van2008visualizing} to project these high-dimensional vectors to two-dimensional space.
Figure~\ref{fig:tsne-tl-example} visualizes an example query. In English, the query is \textit{``What is the schedule predicted for the European single currency?''} The parallel Tagalog query is \textit{``Ano ang hinuhulaang iskedyul para sa iisang uri ng pera sa Europa?''}
Figure~\ref{fig:tsne-tl-overview} provides an overview of all test queries in both English and Tagalog. 
We can see that when used in zero-shot setting, the English and Tagalog token representations generated by teacher model have a clear language boundary. Although starting from a multilingual pre-trained language model, mColBERT is only trained on MS MARCO English data so that only English tokens have the knowledge of query document matching.  Therefore, we observe a large retrieval performance gap on the same model between the English query (i.e., Translate-Test) and the query in the low-resource languages (i.e., mColBERT).
On the other hand, Tagalog token representations generated by the student encoder are much closer to English token representations. More importantly, by reducing the transportation cost of the parallel sentences, \ModelAbbr{} can push Tagalog words toward their corresponding English words. 
Eventually, Tagalog words that are close to their English translations can also obtain retrieval knowledge because the English token representations are generated by the teacher model.
This matches the design purpose of \ModelAbbr{}.

\textbf{Student query encoder in high-resource language.} We repeat the same analysis but for French queries. Figure~\ref{fig:tsne-fr-example} shows the t-SNE visualization of the same query in French. Figure~\ref{fig:tsne-fr-overview} provides the overview of all test queries in both English and French. Different from low-resource languages, we can see that English and French words are already mixed in the representation space so that there is no clear language boundary. Tokens in French are already close to their translations in English. This explains why the effect of knowledge distillation from \ModelAbbr{} is more significant Tagalog than it is in French.

\subsection{Effect of Bitext Data Size}
\begin{figure}[t]
    \captionsetup[subfigure]{font=footnotesize,labelfont=footnotesize}
    \begin{subfigure}[t]{0.45\textwidth}
        \centering
        \includegraphics[width=\linewidth]{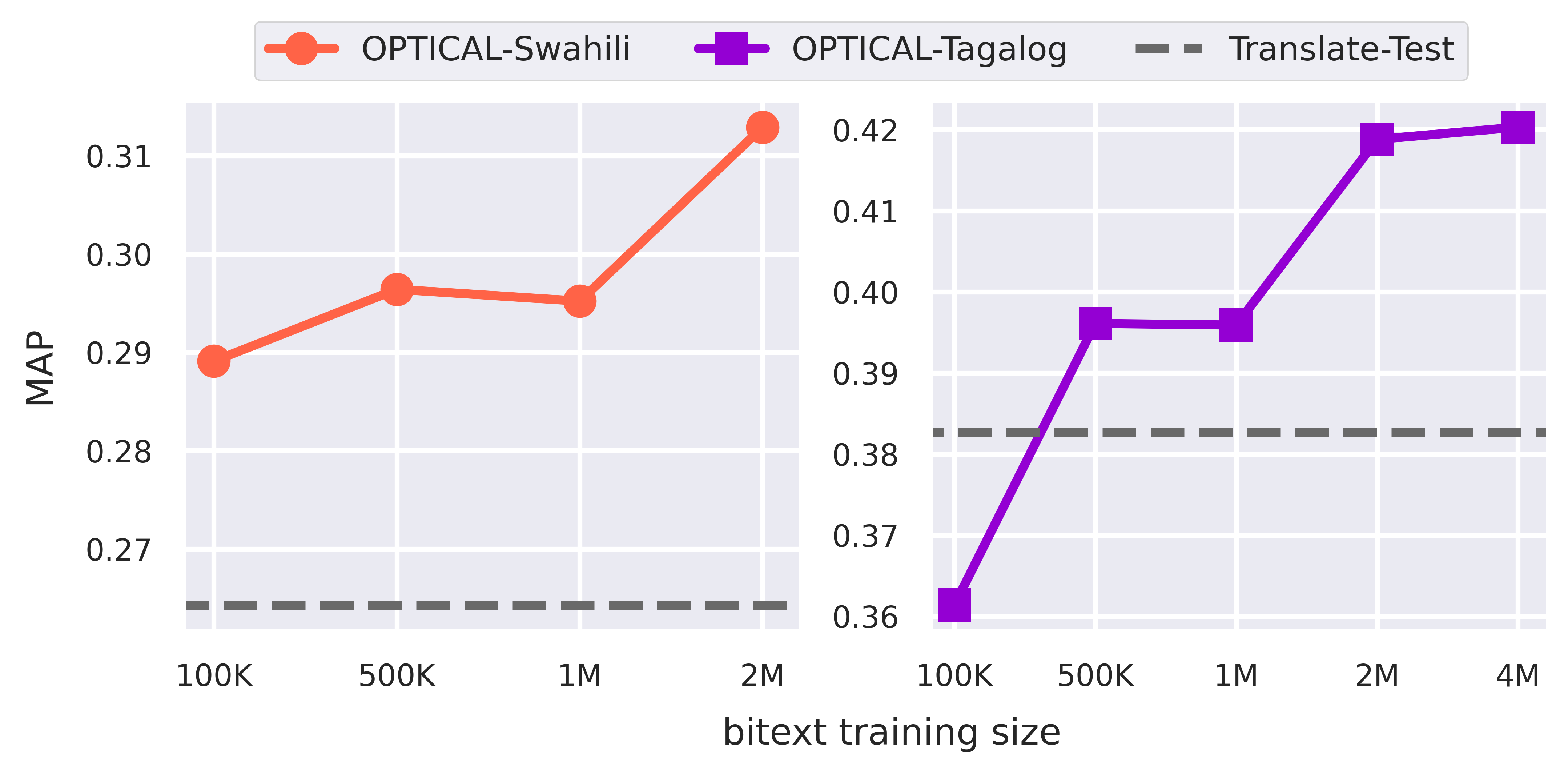}
    \end{subfigure}
\vspace{-12pt}
\caption{Performance with respect to bitext data size.}
\label{fig:bitext}
\end{figure}

\ModelAbbr{} results in Table~\ref{tab:main} are based on a maximum of 2M bitext data used for training distillation. In this experiment, we study the effect of bitext data size on \ModelAbbr{}. We select two low-resource languages with a relatively larger collection of parallel sentences in CCAligned: Swahili (2M) and Tagalog (6.6M). Then we train \ModelAbbr{} using different sizes of the bitext data: 100K, 500K, 1M, 2M, and for Tagalog, we also experiment on 4M bitext data. Figure~\ref{fig:bitext} shows the MAP performance with respect to the size of bitext data. As expected, more bitext data with larger vocabulary size and broader semantic coverage lead to better reranking performance. 
For Swahili, \ModelAbbr{} exceeds the Translate-Test method with a set of only 100K parallel sentences. For Tagalog, starting from 500K, \ModelAbbr{} performs better than Translate-Test. This demonstrates that \ModelAbbr{} is data-efficient.

\subsection{Reduce High-resource to Low-resource}
We hypothesize that the strong performance of the Translate-Test method on high-resource languages is mainly because of the excellent translation quality from the NMT models. Yet a large amount of training data is a prerequisite for the success of NMT. In this experiment, we turn high-resource to low-resource language by limiting the training data size of the NMT models. 
For French and German, we follow the same architecture of the OPUS-MT model but only use subsets with sizes of 5M and 10M pairs, respectively, from the OPUS corpora for training. We compare Translate-Test with the suboptimal NMT models and \ModelAbbr{} in Table~\ref{tab:reduce}. The drops on MAP of Translate-Test indicate the performance of the Translate-Test heavily relies on the NMT model which is data-hungry. Moreover, the knowledge distillation step in \ModelAbbr{} and the training of the NMT model use the same type of data, i.e., the bitext data. Therefore, for learning the translation knowledge in the CLIR task, \ModelAbbr{} is more data efficient than the NMT model.


\begin{table}[t]
    \centering
    \captionsetup{width=\linewidth}
    \caption{Performance comparison of reducing high-resource to low-resource language.}
    \vspace{-5pt}
    \label{tab:reduce}
    \begin{adjustbox}{width=0.45\textwidth}
    \aboverulesep=0ex
    \belowrulesep=0ex
    \renewcommand{\arraystretch}{1.2}
    \begin{tabular}{lcccc}
        \toprule
        \multirow{3}{*}[-6pt]{\textbf{\shortstack[l]{Retrieval\\Methods}}} & \multicolumn{4}{c}{Limited size of NMT model training} \\
        \cmidrule(lr){2-5}
        & \multicolumn{2}{c}{French} & \multicolumn{2}{c}{German} \\
        \cmidrule(lr){2-3} \cmidrule(lr){4-5}
        & 5M & 10M & 5M & 10M \\
        \midrule
        Translate-Test & 0.3820 (-19.8\%) & 0.4525 (-5.0\%) & 0.3971 (-17.8\%) & 0.4667 (-3.4\%)\\
        \midrule
        \ModelAbbr{} (2M) & \multicolumn{2}{c}{\textbf{0.4764}} & \multicolumn{2}{c}{\textbf{0.4832}} \\
        \bottomrule
    \end{tabular}
    \end{adjustbox}
\end{table}

\section{Conclusion}  \label{sec:conclusion}
In this paper, we propose \ModelAbbr{}, an optimal transport knowledge distillation framework for CLIR task involving low-resource languages. \ModelAbbr{} builds CLIR models by separating the retrieval knowledge from the translation knowledge. First, the teacher model learns the retrieval knowledge in a monolingual setting. Then we design a knowledge distillation loss based on the optimal transport costs to transfer the retrieval knowledge to the student model in a cross-lingual setting. 
Because of separating the retrieval and translation knowledge, \ModelAbbr{} greatly reduces the data requirements for building a CLIR model, especially for CLIR tasks involving low-resource languages.
Our comprehensive experimental results show that \ModelAbbr{} significantly outperforms other baselines in low-resource languages, including the NMT models. Further analysis demonstrates the effectiveness and high data efficiency of the knowledge distillation step in \ModelAbbr{}. For future work, we are interested in extending \ModelAbbr{} to transfer other monolingual task-specific knowledge into multilingual space.

\begin{acks}
This research is based upon work supported in part by the Center for Intelligent Information Retrieval, and in part by the Office of the Director of National Intelligence (ODNI), Intelligence Advanced Research Projects Activity (IARPA), via Contract No. 2019-19051600007 under Univ. of Southern California subcontract no. 124338456. The views and conclusions contained herein are those of the authors and should not be interpreted as necessarily representing the official policies, either expressed or implied, of ODNI, IARPA, or the U.S. Government. The U.S. Government is authorized to reproduce and distribute reprints for governmental purposes notwithstanding any copyright annotation therein. Any opinions, findings and conclusions or recommendations expressed in this material are those of the authors and do not necessarily reflect those of the sponsor.
\end{acks}

\bibliographystyle{ACM-Reference-Format}
\bibliography{acmart}
\end{document}